\documentclass[runningheads]{llncs}

\usepackage[mobile]{eccv}
\usepackage{multirow}

\usepackage{eccvabbrv}

\usepackage{graphicx}
\usepackage{booktabs}

\usepackage[accsupp]{axessibility}  

\usepackage[pagebackref,breaklinks,colorlinks]{hyperref}

\usepackage{orcidlink}
\usepackage{subfiles}
\usepackage{mathtools}
\usepackage{wrapfig}
\usepackage[nice]{nicefrac}

\begin{document}

\title{Stylizing Sparse-View 3D Scenes with Hierarchical Neural Representation} 


\author{$\text{Yifan Wang}^*$ \and
$\text{Ang Gao}^*$ \and
Yi Gong \and
Yuan Zeng}

\authorrunning{Yifan Wang et al.}

\institute{Southern University of Science and Technology
}

\maketitle
\def\thefootnote{*}\footnotetext{Authors contributed equally to this work.}

\begin{abstract}

Recently, a surge of 3D style transfer methods has been proposed that leverage the scene reconstruction power of a pre-trained neural radiance field (NeRF). To successfully stylize a scene this way, one must first reconstruct a photo-realistic radiance field from collected images of the scene. However, when only sparse input views are available, pre-trained few-shot NeRFs often suffer from high-frequency artifacts, which are generated as a by-product of high-frequency details for improving reconstruction quality. Is it possible to generate more faithful stylized scenes from sparse inputs by directly optimizing encoding-based scene representation with target style? In this paper, we consider the stylization of sparse-view scenes in terms of disentangling content semantics and style textures. We propose a coarse-to-fine sparse-view scene stylization framework, where a novel hierarchical encoding-based neural representation is designed to generate high-quality stylized scenes directly from implicit scene representations. We also propose a new optimization strategy with content strength annealing to achieve realistic stylization and better content preservation. Extensive experiments demonstrate that our method can achieve high-quality stylization of sparse-view scenes and outperforms fine-tuning-based baselines in terms of stylization quality and efficiency.
\end{abstract}

\section{Introduction}
\label{sec:intro}

\begin{figure}[h]
\centering
\includegraphics[width=1\linewidth]{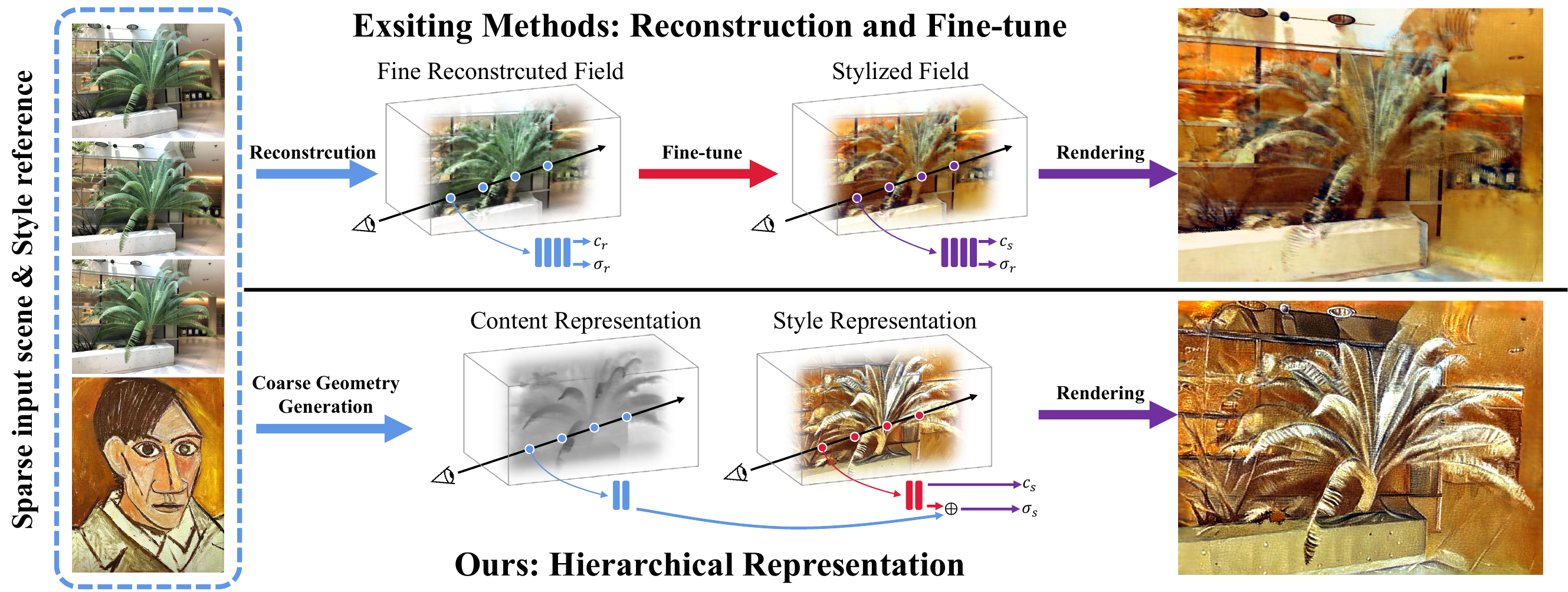}
\caption{The workflow of fine-tuning-based scene stylization method (top) and our coarse-to-fine scene stylization method (bottom). We design a novel hierarchical representation to handle the stylization of sparse-view 3D scenes. In contrast to directly fine-tuning a pre-trained photo-realistic radiance field, we extract low-frequency coarse geometry from sparse inputs as a content component. Then we adopt a followed-up style field to generate stylized appearance and geometry details, where the stylized scene is represented by two fields jointly. Compared with existing fine-tuning-based scene stylization pipelines (e.g., FreeNeRF\ \cite{yang2023freenerf}+ARF\ \cite{zhang2022arf} in our figure), our method yields more reasonable content identity and better style detail.}
\label{fig:first}
\end{figure}

Style transfer is a long-standing problem in computer vision. Scene stylization can edit the appearance of 3D scenes with referenced styles, which opens the way to an intelligent and effective method of digital art and visual design for various virtual reality (VR) and augmented reality (AR) applications. The key challenge of 3D scene stylization lies in rendering a scene from different viewpoints matching the style of a desired image while preserving the semantic content of the scene. This paper focuses on the problem of stylizing sparse-view scenes. Given a few input views of a 3D scene, this paper aims to generate stylized novel views of the scene with multi-view consistency.



Recent 3D scene stylization techniques piggyback on the tremendous success of neural radiance fields (NeRFs)\ \cite{mildenhall2020nerf}, by fine-tuning a pre-trained radiance field so it can generate stylized images from a synthesized novel view. The target stylization is defined with respect to the scene reconstruction, e.g., changing the stylization from a photo-realistic radiance field to an artistic one. When such a relationship holds, the fine-tuning process typically seeks to transform geometric details into high-quality artistic textures. Using this property, many methods demonstrate high-quality scene stylization abilities when a scene is densely observed. Similarly, for stylizing a sparse-view 3D scene, a straightforward solution is to fine tune a pre-trained photo-realistic radiance field reconstructed from sparse input views of the scene. While this method can be used to stylize sparse-view scenes, the final stylization quality depends on how accurately the pre-trained NeRFs reconstruct high-frequency details of the scene.

NeRF works well with multi-view supervision, but it is prone to overfitting to training views and fails to generate correct geometry in few-shot scenarios\ \cite{yu2021pixelnerf, yang2023freenerf}, resulting in unpleasant stylization results. Recently, many few-shot NeRFs have been proposed to overcome these limitations and enhance reconstruction quality. To reconstruct realistic scene from sparse inputs, few-shot NeRFs optimize models to fitting the scene with better high-quality details. This improves the quantitative and qualitative performance, but also brings some high-frequency artifacts as a by-product. Given a pre-trained radiance field reconstructed by a sparse-input NeRF, the fine-tuning-based stylization process has limited capability to re-optimize the high-frequency artifacts into reasonable style textures, as shown in Figure.\ \ref{fig:first}.

In few-shot scenarios, NeRF using low-frequency encoding could produce significantly better semantic representations than those using high-frequency encoding\ \cite{yang2023freenerf, jain2021putting}. Integrating coarse-to-fine radiance field optimization into the style transfer process can potentially improve 3D scene stylization efficiency and final visual performance. In this work, we draw inspiration from this insight and design a novel sparse-view scene stylization framework. We follow the basic principle of ideal style transfer and formulate the scene stylization as a coarse-geometry-to-fine-stylization process, where low-frequency representations can provide a reasonable semantic content of the scene and high-frequency representations are optimized to generate stylized texture. Specifically, we propose a hierarchical encoding-based neural representation for coarse-geometry-to-fine-stylization. It first maps low-frequency encoding to coarse scene geometry. As intermediate output, the coarse radiance field is used to assist the model in directly converting the high-frequency encoding into a stylized scene. Our framework helps to significantly eliminate ambiguity in the optimization process, enabling accurate and robust scene stylization with only sparse view inputs. Moreover, to better balance the stylization effect and the content preservation, we design a new training strategy with content strength annealing.



We conduct extensive experiments and demonstrate that our method can effectively transfer artistic styles to complex 3D scenes with sparse inputs, and outperforms state-of-the-art scene stylization methods both quantitatively and qualitatively. In summary, we make the following contributions:
\begin{itemize}
    \item We propose a new 3D scene stylization framework that integrates coarse-to-fine radiance field into scene stylization, thus enabling efficient and high-quality sparse-view scene stylization.
    

    \item Within the proposed framework, we introduce a hierarchical encoding-based scene representation to model a sparse-view scene from coarse to fine, where the coarse-level representation is first optimized to capture the coarse geometry of a scene from sparse inputs, and then the fine-level representation is directly optimized with the target style to generate the final stylized scene. 
    

    \item We design a new optimization strategy with content annealing for fine 3D stylized scene generation. Our model can generate accurate semantic content in the early phase of stylization optimization, and later gradually synthesizes high-quality stylized textures that faithfully match the reference style.
    \end{itemize}
    
\section{Related Work}
\textbf{Image and video style transfer.} Style transfer aims at generating a new image with the aesthetic style of one image and maintaining the content structure of another. Traditional style transfer methods use handcrafted features to simulate styles\ \cite{jacobs2001image, ashikhmin2003fast}. With the development of deep learning, neural networks have been used for style transfer and achieved impressive visual results. Neural style transfer, first introduced in Gatys \emph{et al.}\ \cite{gatys2016image}, can be optimization-based or feed-forward-based. Early optimization-based methods performed an optimization process to iteratively update the output according to the Gram matrix loss and the content loss. Recent work on loss improvements utilizes alternative loss functions to generate stylized images with better semantic consistency and high-frequency style details\ \cite{li2016combining, gu2018arbitrary}. Subsequently, feed-forward-based methods have been studied to transfer the input images using a single forward pass, which speeds up rendering. By leveraging optical flow\ \cite{ruder2018artistic, huang2017real, wang2020consistent1} or aligning intermediate feature representations\ \cite{deng2021arbitrary} to enforce the temporal consistency across frames, neural style transfer has been successfully extended to videos. Although these works have demonstrated impressive stylization performance on images and videos, they are restricted to the given views and cannot render consistent frames in arbitrary views without considering 3D geometry. Our work draws inspiration from this line of work and uses a NNFM loss \cite{zhang2022arf}, but instead of using a constant factor to control style strength, we introduce a content annealing strategy in the optimization procedure. This allows us to render visually pleasant stylization results with better geometric details.

\noindent \textbf{3D style transfer.} 3D style transfer aims to transfer a reference style into a 3D scene with multi-view consistency. Existing explicit expression-based methods use meshes\ \cite{yin20213dstylenet} or point clouds\ \cite{huang2021learning} to represent 3D scenes. Although promising results have been achieved, these methods are difficult to generalize to complex real-world scenes. To deal with this problem, many implicit expression-based methods\ \cite{jiang2020local, genova2020local, park2019deepsdf, mildenhall2020nerf} have been proposed. Among these methods, NeRF\ \cite{mildenhall2020nerf} is one of the main streams, which is highly related to our work. Chiang \emph{et al.}\ \cite{chiang2022stylizing} first adopt NeRF for 3D scene reconstruction and use pre-trained style hypernetworks for appearance stylization. Later, a large number of follow-up works are presented to improve the rendering efficiency\ \cite{liu2023stylerf, nguyen2022snerf}, quality\ \cite{zhang2022arf, huang2022stylizednerf}, controllability\ \cite{pang2023locally, zhang2023ref}, and generalization\ \cite{liu2023stylerf}. Ref-NPR\ \cite{zhang2023ref} utilizes radiance fields with a 2D reference for controllable scene stylization. ARF\ \cite{zhang2022arf} converts a photo-realistic radiance field into a stylized one using a nearest neighbor-based loss. StyleRF \cite{liu2023stylerf} is a zero-shot high-quality 3D scene stylization method that decomposes the style transformation into sampling-invariant consistent transformation and deferred style transformation. These methods can produce high-quality 3D scene stylization results with a pre-trained radiance field reconstructed from multi-view images. In this work, we consider rendering high-quality stylized scenes from sparse input views. Our insight is to extract low-frequency coarse scene geometry from sparse inputs and use it to assist in rendering the stylized novel views from the hash encoding-based fine scene representation.


\noindent\textbf{Neural Radiance Fields.} NeRF\cite{mildenhall2020nerf} employs Multi-Layer Perceptrons (MLPs) to learn the radiance field of a scene and uses differentiable rendering to reconstruct the scene. Due to its simplicity and exceptional rendering quality, it has become a popular representation for various tasks such as novel view synthesis\ \cite{barron2021mip, zhang2020nerf++, martin2021nerf, wang2024hyb}, 3D generation\ \cite{chan2022efficient, poole2022dreamfusion, metzer2023latent}, scene editing\ \cite{wang2022clip, Haque_2023_ICCV, ye2023intrinsicnerf}, video synthesis\ \cite{tian2023mononerf, li2022neural, park2021hypernerf}. Despite the successes, NeRF still requires hundreds of input images to learn high-quality scene representations and fails to render novel views with a few input views. Many works have been proposed to address this problem by exploiting extra models\ \cite{yao2018mvsnet, yu2021pixelnerf, jain2021putting, wynn2023diffusionerf}, and additional supervision\ \cite{niemeyer2022regnerf, deng2022depth, somraj2023simplenerf}. For sparse view settings, these methods can mitigate blurring artifacts and overfitting in NeRF and render novel views with geometric details. The objective of few-shot NeRFs is to minimize the reconstruction loss, yielding high-quality novel view synthesis from sparse inputs of a scene. In contrast, our goal is to generate visual pleasant stylized scene from sparse input views. Instead of using a photo-realistic pre-trained few-shot radiance field as a content prior for scene stylization, we propose a hierarchical encoding-based sparse-view scene stylization framework, which directly optimizes the fine geometry representation into the stylized scene with multi-view consistency. 

\section{Method}
\begin{figure}[h]
\centering
\includegraphics[width=1\linewidth]{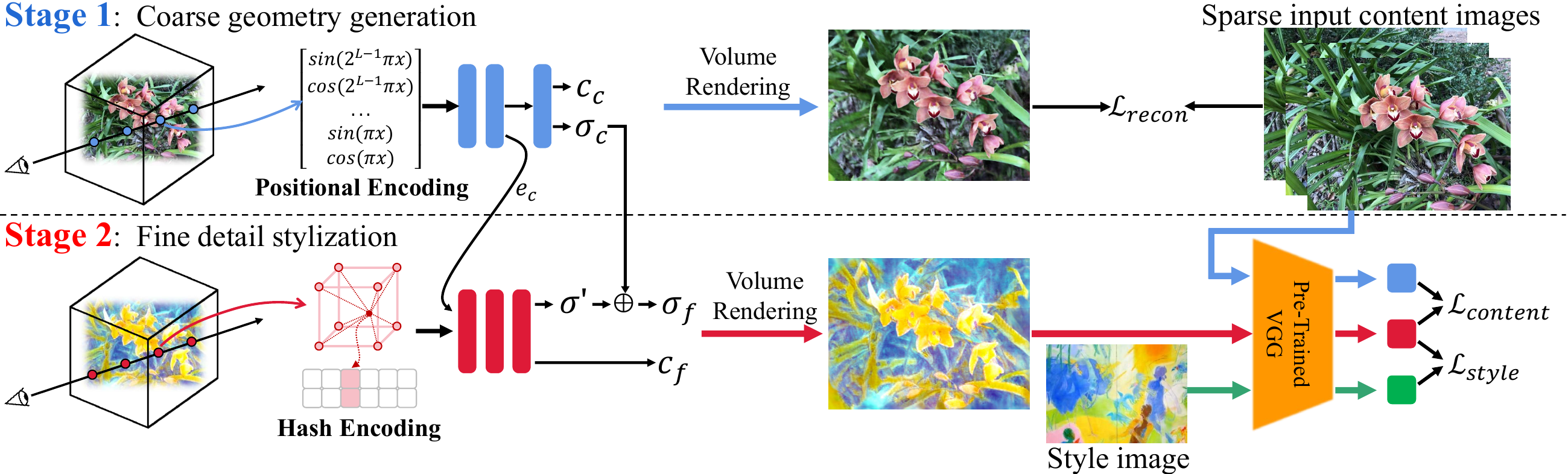}
\caption{Method Overview. We propose a coarse-to-fine framework for sparse-view 3D scene stylization. At the coarse stage, the scene is coarsely generated by mapping low-frequency positional encoding of the 3D position $\mathbf{x}$ and view direction $\mathbf{d}$ to the volume density $\mathbf{\sigma}_{c}$ and RGB color $\mathbf{c}_{c}$. For fine artistic stylization, the second stage models the scene with the multi-resolution dense feature grids and leverages the intermediate implicit features $\mathbf{e}_{c}$ as additional information to assist an MLP decoder in generating high-quality stylized appearance from the feature grids.}
\label{fig:pipeline}
\end{figure}

Fig. \ref{fig:pipeline} shows an overview of our framework. Given a few input images of a 3D scene, our goal is to render arbitrary novel views of the scene with the artistic style of a reference image while maintaining geometry consistency. We formulate this problem as a geometry and appearance-aware optimization problem and introduce a coarse-to-fine framework for high-quality sparse-view scene stylization. We first present the scene geometry and appearance using coarse-to-fine representations (Sec. \ref{sec:coarse}). Then, we optimize the coarse and fine-level representations sequentially for coarse geometry generation and fine stylized scene generation (Sec. \ref{sec:model_training}). 
\subsection{Hierarchical scene representation}
\label{sec:coarse}
 To model a 3D scene with only sparse observations, we propose a hierarchical representation that increases stylization efficiency by generating geometry details proportionally to their expected effect on the final stylization. Instead of using only a pre-trained radiance field to render a photo-realistic scene for stylization, we hierarchically optimize coarse and fine scene representations for coarse scene geometry generation and fine scene stylization. We first adopt a simplified NeRF with low-frequency positional encoding for coarse-level scene generation. Given the output of this coarse representation, we then generate fine stylized novel views with a multi-resolution feature grid-based fine-level scene representation. 
 
\noindent\textbf{Coarse-level representation.}
 To model the coarse scene geometry and layout with few-shot observations, we utilize NeRF, which defines a continuous volumetric field as implicit functions, parameterized by an MLP $\mathcal{F}_{c}$. Given a 3D position $\mathbf{x}$ and a 2D viewing direction $\mathbf{d}$, we use $\mathcal{F}_{c}$ to obtain its corresponding density $\mathbf{\sigma}_{c}$, a geometric feature $\mathbf{e}_{c}$ and 3D color values $\mathbf{c}_{c}$ as 
\begin{equation}
\label{eq:coarse_representation}
(\mathbf{\sigma}_{c}, \mathbf{e}_{c}, \mathbf{c}_{c})=\mathcal{F}_{c}\left(\gamma(\mathbf{x}), \mathbf{d}\right),
\end{equation}
where $\gamma$ is the fixed positional encoding\ \cite{mildenhall2020nerf, tancik2020fourier} that maps the coordinate values in $\mathbf{x}$ to higher dimension. Scene representations with sinusoidal positional encoding are effective at learning high-frequency functions, which assists the MLP in representing fine details. However, when only sparse views of a scene are available for supervision, radiance fields with these high-frequency representations suffer from overfitting and rendering artifacts. To address this issue, our work draws inspiration from\ \cite{jain2021putting} that a simplified NeRF architecture with a shallower MLP and lower maximum frequency positional encoding can avoid high-frequency artifacts and capture the scene geometry roughly without geometric details. We set the level for positional encoding to $7$ and the number of hidden layers of MLP to $6$. 

\noindent\textbf{Fine-level representation.}
\label{sec:fine} 
 While we can obtain the coarse scene geometry from the coarse-level representation, it is essential to render high-quality stylized scene with fine geometry details. Given a reference style image and the coarse geometry representations of a scene, our second stage aims to render stylized novel views of the scene with multi-view consistency. To achieve this, we further model the high-frequency geometric details as residual values using multi-resolution hash-based feature grids\ \cite{muller2022instant} and an MLP decoder. We adopt multi-resolution hash-based feature grids $\mathbf{\psi}(\mathbf{x})$ with $M$ resolution levels. Specifically, the hash encoding of the position $\mathbf{x}$ at the $m$-th level is tri-linearly interpolated features at $8$ corners of its surrounding voxel in a feature grid with resolutions $N_{m}$. $N_{m}$ can be computed as: $N_m:=\left\lfloor N_{\min } \cdot b^l\right\rfloor$, and $b:=\exp \left(\frac{\ln N_{\max }-\ln N_{\min }}{M-1}\right)$, where $N_{min}$ and $N_{max}$ are the lowest and highest resolution, respectively. Here, we consider $N_{min}=128$, $N_{max}=512$, and $M=8$ resolution levels. For each resolution level, the feature dimension is $4$ and the hash table length is $2^{19}$. Then we can predict the residual volume density and stylized color values for $\mathbf{x}$ at fine level by inputting the multi-resolution feature grids and the coarse-level intermediate geometric feature $\mathbf{e}_{c}$ to an MLP $\mathcal{F}_{f}$ with $2$ hidden layers of size $256$:
\begin{equation}
(\mathbf{\sigma}^{\prime},\mathbf{c}_{f})=\mathcal{F}_{f}\left(\mathbf{\psi}(\mathbf{x}),\mathbf{e}_{c}\right),
\end{equation}
where $\mathbf{\sigma}^{\prime}$ is the residual density, and $\mathbf{c}_{f}$ is the predicted color. The geometric features $\mathbf{e}_{c}$ extracted from the coarse-level representation are combined with the multi-resolution feature grids $\mathbf{\psi}(\mathbf{x})$ to assist the MLP in transferring the high-frequency information while preserving the semantic content and overall spatial structure of the scene. The final predicted density $\mathbf{\sigma}_{f}$ is the sum of the coarse-level base density value $\mathbf{\sigma}_{c}$ and the predicted residual density $\mathbf{\sigma}^{\prime}$:
\begin{equation}
\mathbf{\sigma}_{f}=\mathbf{\sigma}^{\prime} + \mathbf{\sigma}_{c}.
\end{equation}
\subsection{Model training}
\label{sec:model_training}
The optimization of our model contains two stages: the first stage aims to optimize the coarse-level representation for generating coarse scene geometry from few-shot input views, and the fine-level representation for fine scene stylization is optimized in the second stage. We also introduce an optimization strategy for annealing content strength during training. 
 
\noindent\textbf{Coarse geometry generation.} We first learn the coarse geometry representation of the target scene from the sparse input images. As in NeRF\ \cite{mildenhall2020nerf}, the expected rendering color $\hat{C}_{c}(\mathbf{r})$ along camera ray $\mathbf{r}$ can be obtained by aggregating the predicted color $\mathbf{c}_{c}$ and density $\mathbf{\sigma}_{c}$ in Eq. (\ref{eq:coarse_representation}) using differentiable volume learning. The coarse scene representation is then optimized with a RGB reconstruction loss as:
\begin{equation}
\mathcal{L}_{recon} =\Vert C_c(\mathbf{r})-\hat{C}_{c}(\mathbf{r}) \Vert^{2}_{2},
\end{equation}
where $C_c(\mathbf{r})$ is the ground truth color. 

\noindent\textbf{Fine stylized scene rendering.} The fine-level scene representation is trained with a decoupled joint loss for artistic appearance generation. The joint loss includes content loss $\mathcal{L}_{content}$ and style loss $\mathcal{L}_{style}$. The former aims to preserve the image-specific features of content images, and the latter constrains the image-specific features to match the style of the reference image. To allow our loss functions to better measure consistency differences across multiple viewpoints, we use the same loss as\ \cite{zhang2022arf} where the content loss $\mathcal{L}_{content}$ computes $\ell_{\text{2}}$ distance between feature maps of rendered and content images to penalize differences on content, and the style loss $\mathcal{L}_{style}$ minimizes the cosine distance between neural features of the rendered image and their corresponding nearest neighboring neural features of the reference style image. We use the $\text{Conv}_3$ block of the pre-trained VGG-16 \cite{simonyan2014very} to extract feature maps. The overall loss function is defined as:
\begin{equation}
\mathcal{L}_{total} =\lambda \mathcal{L}_{content} + \mathcal{L}_{style},
\end{equation}
where $\lambda$ is a weighting factor. 

\noindent\textbf{Annealing content strength.} The weighting factor $\lambda$ in the overall objective loss is to balance the content preservation and stylization effect. Training the model with a larger $\lambda$ leads to better content preservation, while using a smaller $\lambda$ will result in over-stylized images that match the appearance of the style image but show hardly any local details of content images. However, setting $\lambda$ to an appropriate constant value without considering how the training process fits the objective function is a lazy learning strategy and not efficient for generating visually pleasant stylization results. We find neural network-based stylized models often fit the objective function from low to high frequencies during the training process. Annealing content strength can assist the model in capturing more low-frequency details in the early stages of training and smoothly match the style image better as iteration steps of training increase. During training, instead of using a fixed $\lambda$, we smoothly decrease $\lambda$ from a large value $10$ to a small value $0.1$ with increasing the training iterations:  
\begin{equation}
\lambda = 
\begin{cases}
  \lambda_{0} \cdot \alpha^{\frac{t}{T}} & t \leq T \\
  \lambda_{0} \cdot \alpha &  t > T,
\end{cases}
\end{equation}
where $\lambda_{0}$ is the initial value of $\lambda$, $\alpha$ is the decay factor, and $t$ is the iteration index. The $\lambda$ decreases with training iterations when the iteration step is less than $T$ and remains constant thereafter. At the early stage of training, a larger $\lambda$ makes the model focus on learning low-frequency details of the content images. With decreasing $\lambda$ within $T$ iterations exponentially, the goal of model optimization gradually shifts to style matching. The proportion of style loss is increased to a larger constant after $T$ iterations, resulting in a higher emphasis on rendering scenes that visually match the reference style.

\noindent\textbf{Optimization details.} We first use the Adam optimizer\ \cite{kingma2014adam} to optimize our coarse-level scene representation on few-shot conditions. The model is trained for $50K$ iterations with a learning rate of $5e-4$. For training the second stylization stage, we use the Adam optimizer with $\beta_{1}=0.9$ and $\beta_{2}=0.999$ and set the initial learning rate to $5e-3$, which decays with a factor of $0.33$ at $50$th and $100$th iteration. For content strength annealing, we set $\lambda_{0}=10$, $\alpha=0.01$, and $T=100$.

\section{Experiments}
\noindent\textbf{Datasets.} We conduct experiments on the real-world multi-view dataset LLFF\ \cite{mildenhall2019local} and synthetic dataset Blender\ \cite{mildenhall2020nerf} under few-shot settings. LLFF  contains $8$ complex forward-facing scenes. We use a subset of the provided scenes (\textit{Fern}, \textit{Flower}, \textit{Horns}, \textit{Orchids}, \textit{Room}, \textit{and Trex}). For sparse inputs, we follow RegNeRF\ \cite{niemeyer2022regnerf} to select 3 views per scene for training. For Blender dataset, each scene contains multi-view images of an object (\textit{Chair}, \textit{Drums}, \textit{Ficus}, \textit{Hotdog}, \textit{Lego}, \textit{Materials}, \textit{Mic}, \textit{and Ship}). We follow DietNeRF \cite{jain2021putting} to train on $8$ views. To evaluate the stylization effect of our method, we use a diverse set of style images from the WikiArt dataset\ \cite{wikiart} as our reference styles. 
\begin{figure}[t!]
    \centering 
    \includegraphics[width=1\linewidth]{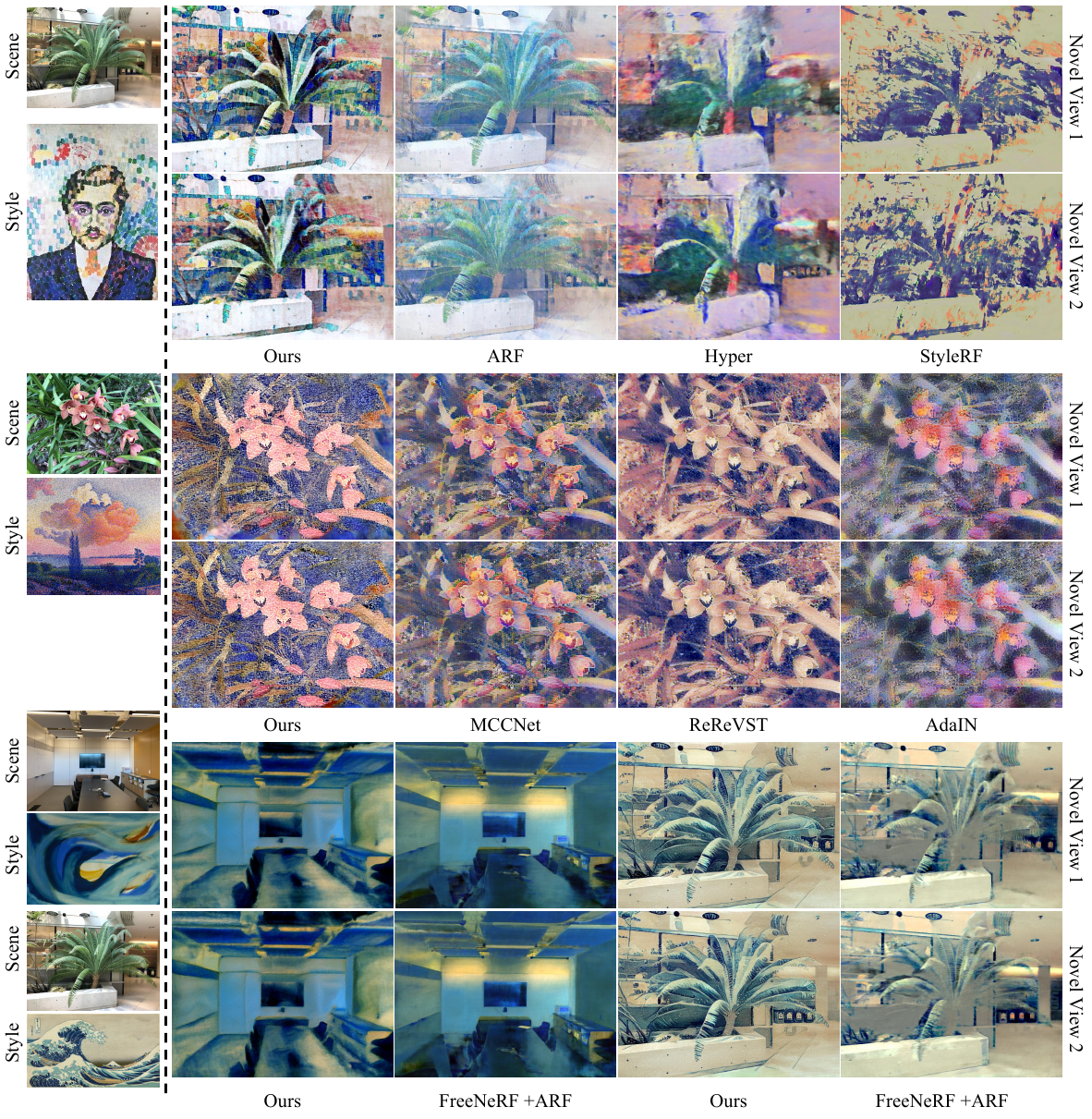}
    \caption{Qualitative comparison of our method with six state-of-the-art style transfer methods ARF \cite{zhang2022arf},  Hyper \cite{chiang2022stylizing}, StyleRF \cite{liu2023stylerf}, MCCNet \cite{deng2021arbitrary}, ReReVST \cite{wang2020consistent1} and AdaIN \cite{huang2017arbitrary} and FreeNeRF\ \cite{yang2023freenerf}+ARF. We render stylized views for all models trained on 3 views. For each of the two sample scenes and reference styles, our method produces clearly better stylized novel views with significantly more faithful style details. Zoom in for better visualization.}
    \label{fig:qualitative}
\end{figure}

\noindent\textbf{Baselines.} We compare against the state-of-the-art scene stylization methods, including three NeRF-based 3D style transfer methods (ARF \cite{zhang2022arf}, StyleRF \cite{liu2023stylerf} and Hyper \cite{chiang2022stylizing}), a 2D style transfer method (AdaIN \cite{huang2017arbitrary}), two video style transfer methods (MCCNet \cite{deng2021arbitrary}, ReReVST \cite{wang2020consistent1}). For the NeRF-based baselines, we use their official code and the same sparse inputs as our method. For AdaIN, MCCNet, and ReReVST, since these stylization methods require multi-view images as input, we use our coarse scene representation as content input and their official pre-trained models for stylization.  

\subsection{Comparisons}

\begin{figure}[h]
   \centering
   \includegraphics[width=1\linewidth]{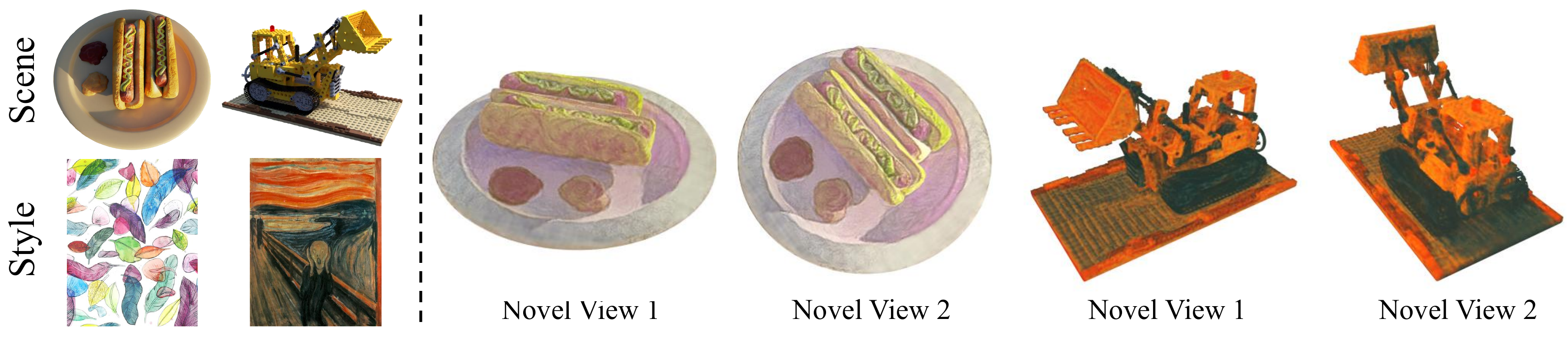}
   \caption{More 3D scene stylization results on Blender dataset. Our method can generate high-quality stylization images with multi-view consistency.}
   \label{fig:figure2}
\end{figure}

    \noindent\textbf{Qualitative comparison.} Fig. \ref{fig:qualitative} shows qualitative comparisons with baselines. It shows that our method generates cross-view consistent 3D stylization results with a better style match to the reference image and with more precise geometric details. 
    3D-based baselines ARF, Hyper, and StyleRF fail to capture fine-level geometric details as well as the style of the reference image. This can be explained by the fact that the three methods are fine-tuning-based methods, and the final stylization performance relies on the scene reconstruction quality obtained from multi-view supervision-based NeRFs. Although FreeNeRF$+$ARF can reconstruct better overall geometry and content identity than 3D-based baselines, the geometric details of small objects are blurry and stylized effects are inadequate. The image style transfer method AdaIN yields more inconsistency artifacts than other methods, since each frame is stylized independently. Video-based methods MCCNet and ReReVST produce less flicking results than AdaIN, since they consider the temporal consistency in stylization optimization. However, they still fail to preserve the consistency between two far-away viewpoints and suffer from texture-sticking artifacts when shifting the view. Our method benefits from both hierarchical scene representation and content strength annealing. It effectively captures coarse scene geometry to assist the feature grid-based model in rendering better stylized views. Fig. \ref{fig:figure2} shows the scene stylization results of our method on Blender \cite{mildenhall2020nerf}. For a more thorough comparison, the rendered views and videos are offered in supplementary materials. 
    
    \noindent\textbf{Quantitative comparison.} We evaluate the consistency between different stylized novel views in terms of the masked RMSE metric, SSIM metric, and LPIPS metric\ \cite{zhang2018unreasonable}. We follow\ \cite{liu2023stylerf} to warp views and calculate the corresponding masked RMSE, SSIM, and LPIPS. Specifically, we use the measurement implemented from\ \cite{nguyen2022snerf, liu2023stylerf} to compute the short-range consistency and far-away views to compute the long-range consistency. Table\  \ref{tab:quantitative} shows that our method outperforms 3D-based and image-based baselines, including ARF, Hyper, StyleRF, FreeNeRF$+$ARF and AdaIN, by a significant margin. Compared with the video-based methods MCCNet and ReReVST, our method achieves Superior consistency. In addition, although the numeric margin between our method and ReReVST\ \cite{wang2020consistent1} is very small, ReReVST fails to capture the desired style of the reference image in some cases, as shown in Fig. \ref{fig:qualitative}.

    \begin{table}[h]
    \caption{Quantitative comparisons on short and long-range consistency.}
      \centering
      \scalebox{0.75}{
      \begin{tabular}{@{}lcccccc@{}}
        \toprule
        \multirow{2}{*}{\textbf{Method}} & \multicolumn{3}{c}{Short-range Consistency} & \multicolumn{3}{c}{Long-range Consistency}\\
        \cmidrule(lr){2-4} \cmidrule(lr){5-7}
        & \textbf{RMSE}($\downarrow$) & \textbf{SSIM}($\uparrow$) & \textbf{LPIPS}($\downarrow$) & \textbf{RMSE}($\downarrow$) & \textbf{SSIM}($\uparrow$) & \textbf{LPIPS}($\downarrow$)\\
        AdaIN \cite{huang2017arbitrary} & 0.189 & 0.618 & 0.276 & 0.231 & 0.601 & 0.295 \\
        MCCNet \cite{deng2021arbitrary} & \colorbox{yellow!40}{0.167} & \colorbox{yellow!40}{0.687} & \colorbox{yellow!40}{0.235} & 0.230 & 0.652 & \colorbox{yellow!40}{0.267}\\
        ReReVST \cite{wang2020consistent1} & \colorbox{orange!40}{0.147} & \colorbox{orange!40}{0.716} & \colorbox{orange!40}{0.211} & \colorbox{yellow!40}{0.180} & \colorbox{yellow!40}{0.685} & \colorbox{orange!40}{0.239} \\
        Hyper \cite{chiang2022stylizing} & 0.233 & 0.655 & 0.325 & 0.240 & \colorbox{red!40}{\textbf{0.699}} & 0.303\\
        ARF \cite{zhang2022arf} & 0.171 & 0.603 & 0.297  & \colorbox{orange!40} {0.178} & 0.651 & 0.279\\
        StyleRF \cite{liu2023stylerf} & 0.287 & 0.539 & 0.378 & 0.318 & 0.587 & 0.369\\
        FreeNeRF \cite{yang2023freenerf}+ARF \cite{zhang2022arf} & 0.206 & 0.685 & 0.269 & 0.233 & 0.674 & 0.273\\
        \textbf{Ours} & \colorbox{red!40}{\textbf{0.137}} & \colorbox{red!40}{\textbf{0.720}} & \colorbox{red!40}{\textbf{0.186}} & \colorbox{red!40}{\textbf{0.176}} & \colorbox{orange!40}{0.686} & \colorbox{red!40}{\textbf{0.225}}\\
        \bottomrule
      \end{tabular}
      }
      
      \label{tab:quantitative}
    \end{table}

\begin{figure}[h]
        \centering
        \includegraphics[width=0.8\linewidth]{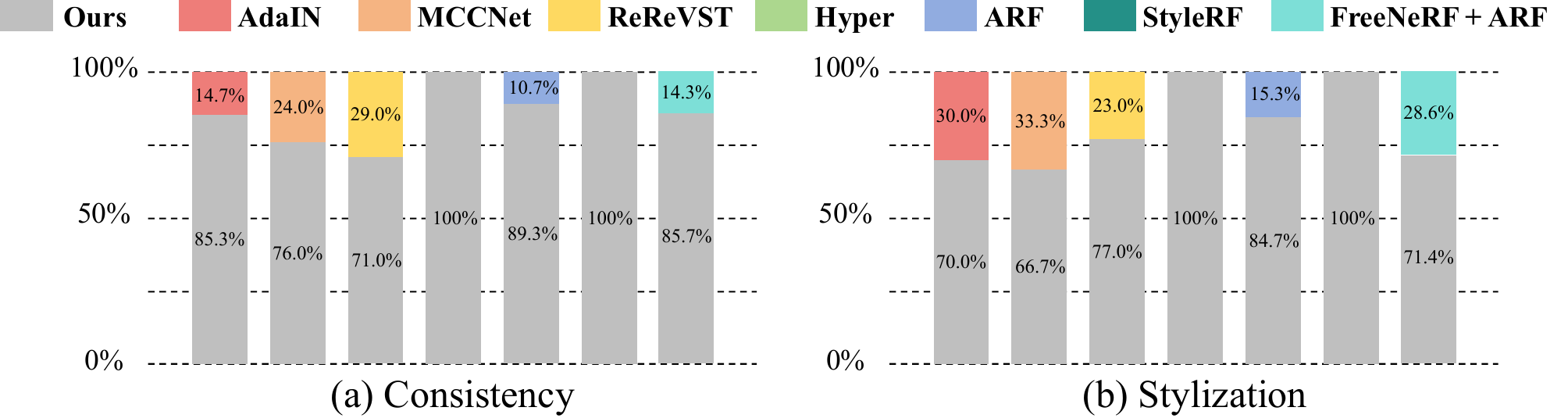}
        \caption{User preference study. We present two videos of stylized results, one generated by our method (gray) and one by another method (other colors). Our method wins more preferences both in the artistic stylization and multi-view consistency.}
        \label{fig:user_study}
    \end{figure}

We conduct a user study to further evaluate the visual performance of the stylization methods. We prepare ten series of stylized views in the LLFF dataset \cite{mildenhall2019local} and invite 30 participants, including 20 males and 10 females, in our study. For each user, we show a reference style image and two videos of the same scene and style, one rendered by our method and the other by a random compared baseline. The user is then asked to select the one that has more consistent content across different views (e.g., less flickering and less texture sticking) and the one that matches the style of a reference style image better. We collect 1800 votes for each evaluating indicator and presented the results in Fig. \ref{fig:user_study}. We observe that our method gets more preference over the other methods in terms of both stylization quality and consistency.  
\subsection{Ablation Studies}
     We do ablation studies to validate the effectiveness of our design choices, including coarse-to-fine framework, hierarchical representation, fine-level geometry $\mathbf{\sigma}_{f}$, and content annealing.
    
\begin{figure}[h]
        \centering
        \includegraphics[width=1\linewidth]{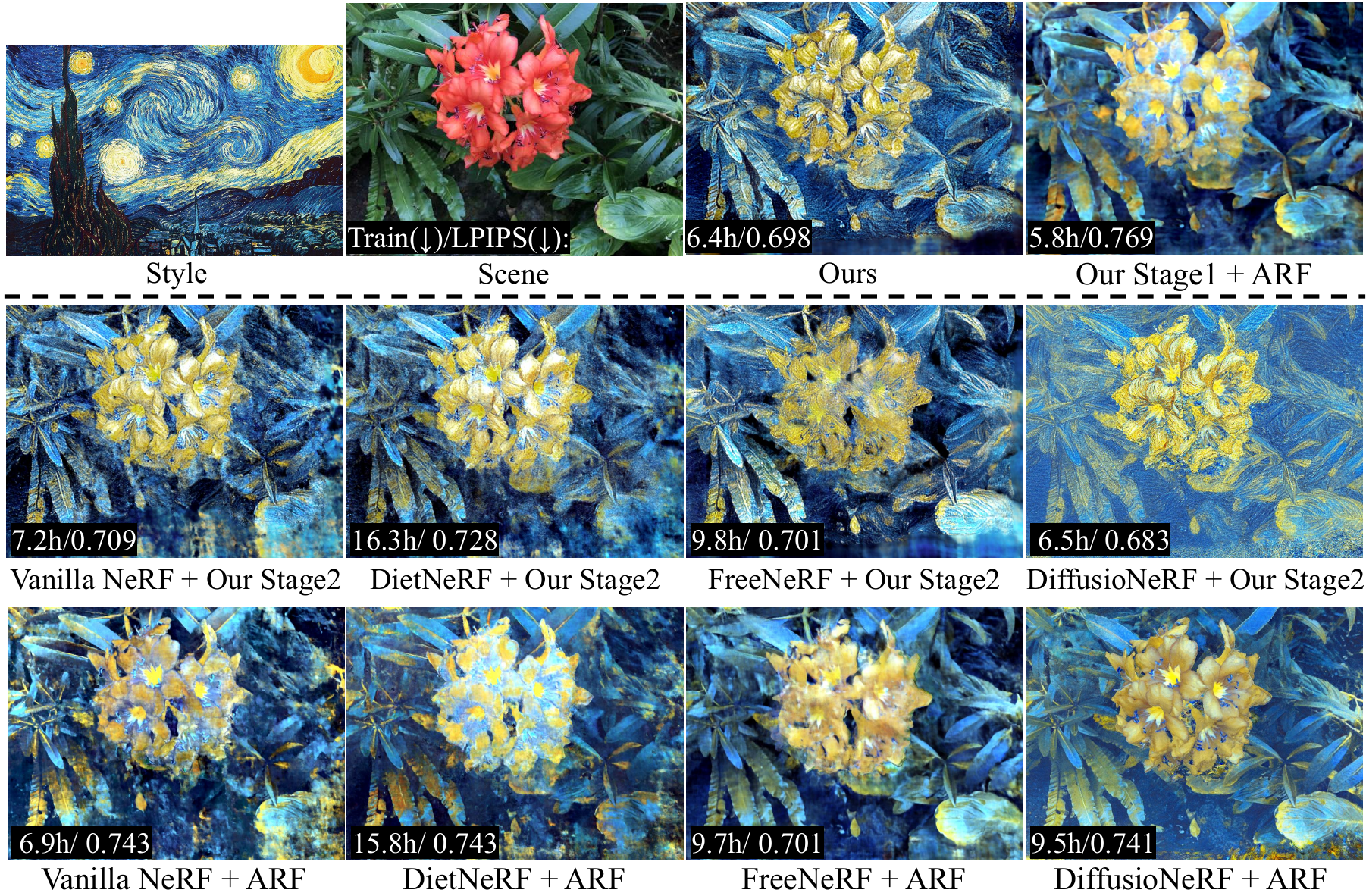}
        \caption{Sparse-view 3D scene stylization using different frameworks and stylization training strategies. The rightmost two images in the top group are stylization results of our method and ARF-based fune-tuning of our coarse radiance field. The top images in the bottom group are generated by replacing our coarse radiance field with different fine radiance fields (vanilla NeRF \cite{mildenhall2020nerf}, DietNeRF \cite{jain2021putting}, FreeNeRF \cite{yang2023freenerf}, DiffusioNeRF \cite{wynn2023diffusionerf}). The bottom row images in the bottom group are rendered by directly fine-tuning different pre-trained fine radiance fields with ARF. Our method is computationally efficient and produces significantly better stylized novel views, while preserving more faithful content identity. We provide more comparison results in the supplementary material.} 
     \label{fig:frequency}
\end{figure}

      \noindent\textbf{Coarse-to-fine framework.} To analyze the stylization effect and efficiency of the proposed coarse-to-fine framework, we compare our method to nine variations. Four variations replace our first stage with finer scene representations reconstructed from vanilla NeRF \cite{mildenhall2020nerf} and few-shot NeRFs (DietNeRF \cite{jain2021putting}, FreeNeRF \cite{yang2023freenerf}, DiffusioNeRF \cite{wynn2023diffusionerf}). The other five variations use ARF\ \cite{zhang2022arf} to fine-tune different pre-trained radiance fields, including our coarse radiance field, vanilla NeRF, DietNeRF, FreeNeRF and DiffusioNeRF. Fig.\ \ref{fig:frequency} shows that the stylization quality of ARF-based fine-tuning methods depends on the quality of the reconstructed field.
      Our coarse radiance field can produce generally accurate geometry for ARF-based scene stylization, but the stylized novel views suffer from blurring artifacts. Vanilla NeRF with high-frequency positional encoding fails to generate correct geometry for scene stylization, resulting in high-frequency artifacts in stylized results. Few-shot NeRF-based methods (DietNeRF/FreeNeRF/DiffusioNeRF+Our stage 2/ARF) achieve better stylized novel views with more accurate geometric details and less high-frequency artifacts. This can be explained by the fact that few-shot NeRFs can produce much better reconstructed geometric details for ARF-based fine-tuning or our fine stylization optimization. In addition, the stylized results in each column show that the fine scene stylization in our second stage can effectively improve the scene stylization quality. Moreover, our method achieves better scene stylization quality than the nine variations. Our method is also computationally more efficient than fine reconstructed radiance field-based variations.

\begin{figure}[h]
        \centering
        \includegraphics[width=1\linewidth]{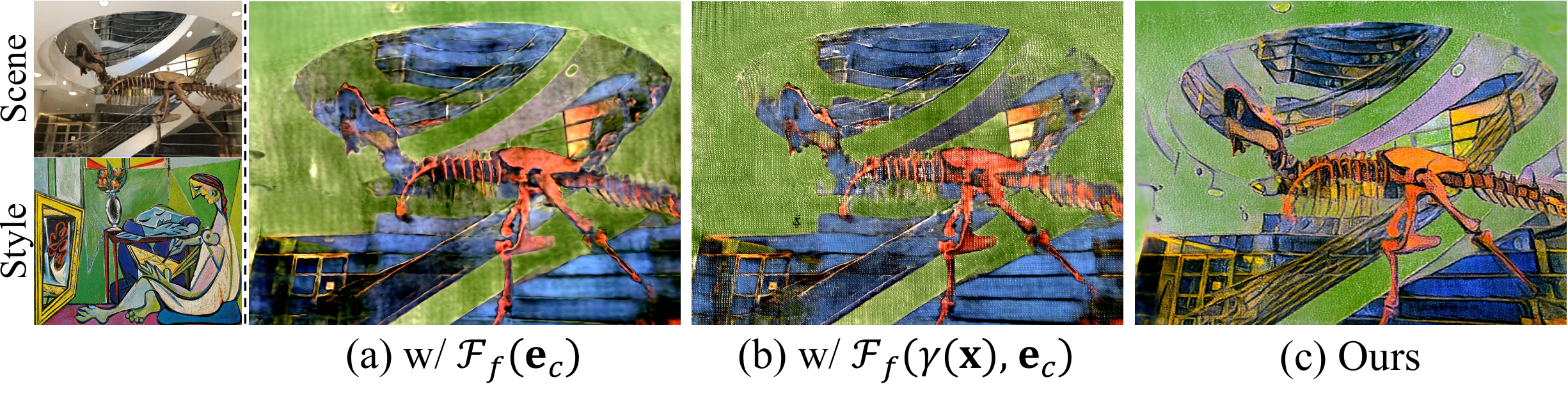}
        \caption{Ablation study on encoding strategy in fine-level representation. (a) shows the fine-level representation with only the latent geometry feature $\mathbf{e}_{c}$. (b) shows the one that replaces the multi-resolution feature grids with a high-frequency positional encoding. (c) shows the result of our design. Our method uses multi-resolution feature grids and latent geometry feature $\mathbf{e}_{c}$ as the input of the stylization MLP produces clearly better scene stylization results with less blurring artifacts.}
        \label{fig:Embedding}
    \end{figure}

     \noindent\textbf{Hierarchical representation.} We compare our method to two variations. The first variation uses the geometry features $\mathbf{e}_{c}$ as the only input in the fine-level representation. This variation leads to blurring effect in the rendered result, as illustrated in Fig. \ref{fig:Embedding} (a). The second one is to use high-frequency positional encoding instead of the multi-resolution hash encoding in the fine-level representation. Fig. \ref{fig:Embedding} (b) shows that this variation causes periodic artifacts, indicating that the multi-resolution feature grid can indeed assist the model in rendering more realistic stylization results with better geometric details. 

\begin{figure}[h]
        \centering
        \includegraphics[width=0.8\linewidth]{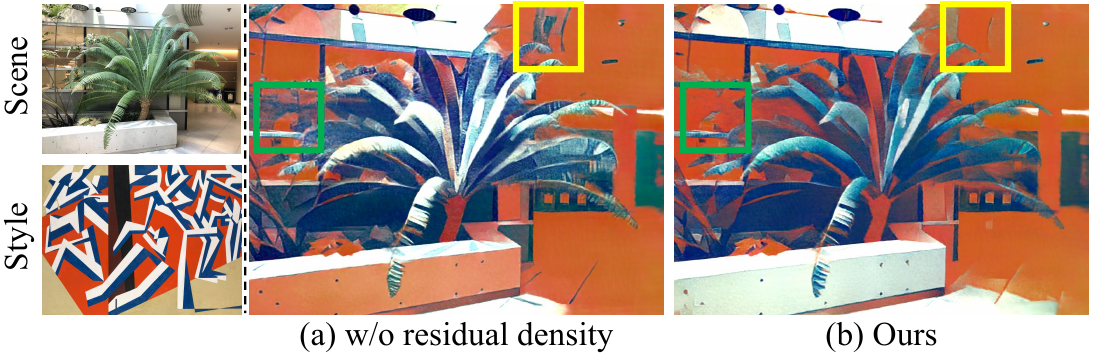}
        \caption{Ablation study on fine-level geometry $\mathbf{\sigma}_{f}$. Our model with residual density in stylization optimization enables the generation of better stylization details that match the reference style. 
        }
        
        \label{fig:density}
    \end{figure}

    \noindent\textbf{Fine-level geometry $\mathbf{\sigma}_{f}$.} To verify the design choices of our fine-level geometry $\mathbf{\sigma}_{f}$, we keep the architecture identical but only use the output density $\mathbf{\sigma}_{c}$ in equation (\ref{eq:coarse_representation}) to produce the stylization views (\emph{w/o residual density}). Fig. \ref{fig:density} clearly shows that our fine-level geometry $\mathbf{\sigma}_{f}$ with residual density leads to better stylization quality with more stylized details that match the reference style.

\begin{figure}[h]
        \centering
        \includegraphics[width=1\linewidth]{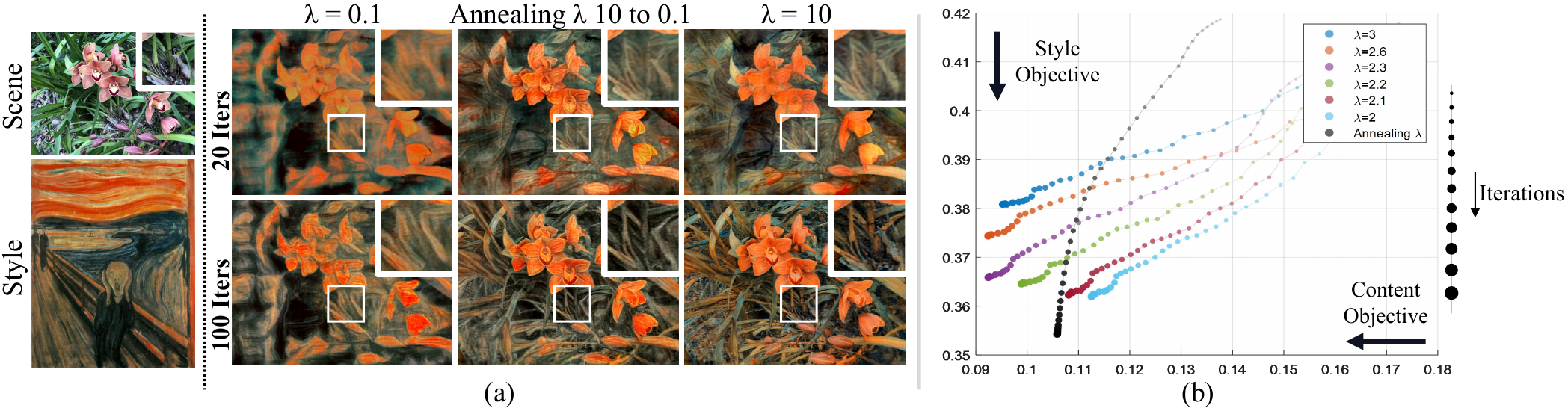}
        \caption{Ablation study on content annealing. (a) shows stylization results of our models with different $\lambda$ trained after 20 and 100 iterations. (b) shows that our content annealing strategy achieves Pareto optimality compared to using various constants $\lambda$. 
        }
        \label{fig:Annealing}
    \end{figure}

\noindent\textbf{Content annealing.} To verify the effectiveness of our content strength annealing on controlling the generation of scene style, we compare our optimization strategy to two variations with different settings of $\lambda$. As shown in Fig. \ref{fig:Annealing} (a), a larger constant factor $\lambda$ enables the stylization field to capture accurate geometry in the early phase of training, but render inferior stylization views that less match the reference style in the late phase of training. In contrast, a small constant $\lambda$ indicates a strong emphasis on style, resulting in limited content matching between the input scene and the stylized one. Our method with content annealing produces clearly better stylization quality while preserving the content structure of the scene. The quantitative comparison results in Fig. \ref{fig:Annealing} (b) illustrate optimization traces of the content loss and style loss during training. Our design achieves Pareto optimality compared to using various constants $\lambda$ for content and style optimization.
\section{Conclusion}
We presented a novel 3D style transfer framework for sparse-view scene stylization, which enables visually pleasant stylized novel view generation. The proposed framework includes a new hierarchical scene representation for directly optimizing the fine-level scene representations into stylized scenes. During the stylization training, a content annealing strategy was introduced to achieve a better balance of the content preservation and the scene stylization effect. We demonstrated that the effectiveness of our design on generating high-quality stylized scene from sparse input views. Experiments on both synthetic and real-world scenes showed that our method achieves superior 3D stylization quality and efficiency over baselines when only sparse views of a scene are available.


%
%
\bibliographystyle{splncs04}
\bibliography{main}

\begin{thebibliography}{10}
\providecommand{\url}[1]{\texttt{#1}}
\providecommand{\urlprefix}{URL }
\providecommand{\doi}[1]{https://doi.org/#1}

\bibitem{ashikhmin2003fast}
Ashikhmin, N.: Fast texture transfer. IEEE computer Graphics and Applications  \textbf{23}(4),  38--43 (2003)

\bibitem{barron2021mip}
Barron, J.T., Mildenhall, B., Tancik, M., Hedman, P., Martin-Brualla, R., Srinivasan, P.P.: {Mip-NeRF}: A multiscale representation for anti-aliasing neural radiance fields. In: ICCV. pp. 5855--5864 (2021)

\bibitem{chan2022efficient}
Chan, E.R., Lin, C.Z., Chan, M.A., Nagano, K., Pan, B., De~Mello, S., Gallo, O., Guibas, L.J., Tremblay, J., Khamis, S., et~al.: Efficient geometry-aware 3d generative adversarial networks. In: CVPR. pp. 16123--16133 (2022)

\bibitem{chiang2022stylizing}
Chiang, P.Z., Tsai, M.S., Tseng, H.Y., Lai, W.S., Chiu, W.C.: Stylizing 3d scene via implicit representation and hypernetwork. In: WACV. pp. 1475--1484 (2022)

\bibitem{deng2022depth}
Deng, K., Liu, A., Zhu, J.Y., Ramanan, D.: {Depth-supervised NeRF}: Fewer views and faster training for free. In: CVPR. pp. 12882--12891 (2022)

\bibitem{deng2021arbitrary}
Deng, Y., Tang, F., Dong, W., Huang, H., Ma, C., Xu, C.: Arbitrary video style transfer via multi-channel correlation. In: AAAI. vol.~35, pp. 1210--1217 (2021)

\bibitem{gatys2016image}
Gatys, L.A., Ecker, A.S., Bethge, M.: Image style transfer using convolutional neural networks. In: CVPR. pp. 2414--2423 (2016)

\bibitem{genova2020local}
Genova, K., Cole, F., Sud, A., Sarna, A., Funkhouser, T.: Local deep implicit functions for 3d shape. In: CVPR. pp. 4857--4866 (2020)

\bibitem{gu2018arbitrary}
Gu, S., Chen, C., Liao, J., Yuan, L.: Arbitrary style transfer with deep feature reshuffle. In: CVPR. pp. 8222--8231 (2018)

\bibitem{Haque_2023_ICCV}
Haque, A., Tancik, M., Efros, A.A., Holynski, A., Kanazawa, A.: {Instruct-NeRF2NeRF}: Editing 3d scenes with instructions. In: ICCV. pp. 19740--19750 (October 2023)

\bibitem{huang2017real}
Huang, H., Wang, H., Luo, W., Ma, L., Jiang, W., Zhu, X., Li, Z., Liu, W.: Real-time neural style transfer for videos. In: CVPR. pp. 783--791 (2017)

\bibitem{huang2021learning}
Huang, H.P., Tseng, H.Y., Saini, S., Singh, M., Yang, M.H.: Learning to stylize novel views. In: ICCV. pp. 13869--13878 (2021)

\bibitem{huang2017arbitrary}
Huang, X., Belongie, S.: Arbitrary style transfer in real-time with adaptive instance normalization. In: ICCV. pp. 1501--1510 (2017)

\bibitem{huang2022stylizednerf}
Huang, Y.H., He, Y., Yuan, Y.J., Lai, Y.K., Gao, L.: {StylizedNeRF}: consistent 3d scene stylization as stylized nerf via 2d-3d mutual learning. In: CVPR. pp. 18342--18352 (2022)

\bibitem{jacobs2001image}
Jacobs, C., Salesin, D., Oliver, N., Hertzmann, A., Curless, A.: Image analogies. In: Proceedings of Siggraph. pp. 327--340 (2001)

\bibitem{jain2021putting}
Jain, A., Tancik, M., Abbeel, P.: {Putting NeRF on a diet}: Semantically consistent few-shot view synthesis. In: ICCV. pp. 5885--5894 (2021)

\bibitem{jiang2020local}
Jiang, C., Sud, A., Makadia, A., Huang, J., Nie{\ss}ner, M., Funkhouser, T., et~al.: Local implicit grid representations for 3d scenes. In: CVPR. pp. 6001--6010 (2020)

\bibitem{kingma2014adam}
Kingma, D.P., Ba, J.: Adam: A method for stochastic optimization. arXiv preprint arXiv:1412.6980  (2014)

\bibitem{li2016combining}
Li, C., Wand, M.: Combining markov random fields and convolutional neural networks for image synthesis. In: CVPR. pp. 2479--2486 (2016)

\bibitem{li2022neural}
Li, T., Slavcheva, M., Zollhoefer, M., Green, S., Lassner, C., Kim, C., Schmidt, T., Lovegrove, S., Goesele, M., Newcombe, R., et~al.: Neural 3d video synthesis from multi-view video. In: CVPR. pp. 5521--5531 (2022)

\bibitem{liu2023stylerf}
Liu, K., Zhan, F., Chen, Y., Zhang, J., Yu, Y., El~Saddik, A., Lu, S., Xing, E.P.: {StyleRF}: Zero-shot 3d style transfer of neural radiance fields. In: CVPR. pp. 8338--8348 (2023)

\bibitem{martin2021nerf}
Martin-Brualla, R., Radwan, N., Sajjadi, M.S., Barron, J.T., Dosovitskiy, A., Duckworth, D.: {NeRF in the wild}: Neural radiance fields for unconstrained photo collections. In: CVPR. pp. 7210--7219 (2021)

\bibitem{metzer2023latent}
Metzer, G., Richardson, E., Patashnik, O., Giryes, R., Cohen-Or, D.: {Latent-NeRF} for shape-guided generation of 3d shapes and textures. In: CVPR. pp. 12663--12673 (2023)

\bibitem{mildenhall2019local}
Mildenhall, B., Srinivasan, P.P., Ortiz-Cayon, R., Kalantari, N.K., Ramamoorthi, R., Ng, R., Kar, A.: Local light field fusion: Practical view synthesis with prescriptive sampling guidelines. ACM TOG  \textbf{38}(4),  1--14 (2019)

\bibitem{mildenhall2020nerf}
Mildenhall, B., Srinivasan, P.P., Tancik, M., Barron, J.T., Ramamoorthi, R., Ng, R.: {NeRF}: Representing scenes as neural radiance fields for view synthesis. In: ECCV. pp. 405--421. Springer (2020)

\bibitem{muller2022instant}
M{\"u}ller, T., Evans, A., Schied, C., Keller, A.: Instant neural graphics primitives with a multiresolution hash encoding. ACM ToG  \textbf{41}(4),  1--15 (2022)

\bibitem{nguyen2022snerf}
Nguyen-Phuoc, T., Liu, F., Xiao, L.: {SNeRF}: stylized neural implicit representations for 3d scenes. arXiv preprint arXiv:2207.02363  (2022)

\bibitem{wikiart}
Nichol, K.: Painter by numbers, wikiart. (2016), \url{https://www.kaggle.com/competitions/painter-by-numbers}

\bibitem{niemeyer2022regnerf}
Niemeyer, M., Barron, J.T., Mildenhall, B., Sajjadi, M.S., Geiger, A., Radwan, N.: {RegNeRF}: Regularizing neural radiance fields for view synthesis from sparse inputs. In: CVPR. pp. 5480--5490 (2022)

\bibitem{pang2023locally}
Pang, H.W., Hua, B.S., Yeung, S.K.: Locally stylized neural radiance fields. In: ICCV. pp. 307--316 (2023)

\bibitem{park2019deepsdf}
Park, J.J., Florence, P., Straub, J., Newcombe, R., Lovegrove, S.: {DeepSDF}: Learning continuous signed distance functions for shape representation. In: CVPR. pp. 165--174 (2019)

\bibitem{park2021hypernerf}
Park, K., Sinha, U., Hedman, P., Barron, J.T., Bouaziz, S., Goldman, D.B., Martin-Brualla, R., Seitz, S.M.: {HyperNeRF}: A higher-dimensional representation for topologically varying neural radiance fields. arXiv preprint arXiv:2106.13228  (2021)

\bibitem{poole2022dreamfusion}
Poole, B., Jain, A., Barron, J.T., Mildenhall, B.: {DreamFusion}: Text-to-3d using 2d diffusion. arXiv preprint arXiv:2209.14988  (2022)

\bibitem{ruder2018artistic}
Ruder, M., Dosovitskiy, A., Brox, T.: Artistic style transfer for videos and spherical images. IJCV  \textbf{126}(11),  1199--1219 (2018)

\bibitem{simonyan2014very}
Simonyan, K., Zisserman, A.: Very deep convolutional networks for large-scale image recognition. arXiv preprint arXiv:1409.1556  (2014)

\bibitem{somraj2023simplenerf}
Somraj, N., Karanayil, A., Soundararajan, R.: {SimpleNeRF}: Regularizing sparse input neural radiance fields with simpler solutions. In: SIGGRAPH Asia 2023 Conference Papers. pp. 1--11 (2023)

\bibitem{tancik2020fourier}
Tancik, M., Srinivasan, P., Mildenhall, B., Fridovich-Keil, S., Raghavan, N., Singhal, U., Ramamoorthi, R., Barron, J., Ng, R.: Fourier features let networks learn high frequency functions in low dimensional domains. NeurIPS  \textbf{33},  7537--7547 (2020)

\bibitem{tian2023mononerf}
Tian, F., Du, S., Duan, Y.: {MonoNeRF}: Learning a generalizable dynamic radiance field from monocular videos. In: ICCV. pp. 17903--17913 (2023)

\bibitem{wang2022clip}
Wang, C., Chai, M., He, M., Chen, D., Liao, J.: {CLIP-NeRF}: Text-and-image driven manipulation of neural radiance fields. In: CVPR. pp. 3835--3844 (2022)

\bibitem{wang2020consistent1}
Wang, W., Yang, S., Xu, J., Liu, J.: Consistent video style transfer via relaxation and regularization. IEEE TIP  \textbf{29},  9125--9139 (2020)

\bibitem{wang2024hyb}
Wang, Y., Gong, Y., Zeng, Y.: {Hyb-NeRF}: A multiresolution hybrid encoding for neural radiance fields. In: WACV. pp. 3689--3698 (2024)

\bibitem{wynn2023diffusionerf}
Wynn, J., Turmukhambetov, D.: {DiffusioNeRF}: Regularizing neural radiance fields with denoising diffusion models. In: CVPR. pp. 4180--4189 (2023)

\bibitem{yang2023freenerf}
Yang, J., Pavone, M., Wang, Y.: {FreeNeRF}: Improving few-shot neural rendering with free frequency regularization. In: CVPR. pp. 8254--8263 (2023)

\bibitem{yao2018mvsnet}
Yao, Y., Luo, Z., Li, S., Fang, T., Quan, L.: {MVSNet}: Depth inference for unstructured multi-view stereo. In: ECCV. pp. 767--783 (2018)

\bibitem{ye2023intrinsicnerf}
Ye, W., Chen, S., Bao, C., Bao, H., Pollefeys, M., Cui, Z., Zhang, G.: {IntrinsicNeRF}: Learning intrinsic neural radiance fields for editable novel view synthesis. In: ICCV. pp. 339--351 (2023)

\bibitem{yin20213dstylenet}
Yin, K., Gao, J., Shugrina, M., Khamis, S., Fidler, S.: {3DStyleNet}: Creating 3d shapes with geometric and texture style variations. In: ICCV. pp. 12456--12465 (2021)

\bibitem{yu2021pixelnerf}
Yu, A., Ye, V., Tancik, M., Kanazawa, A.: {pixelNeRF}: Neural radiance fields from one or few images. In: CVPR. pp. 4578--4587 (2021)

\bibitem{zhang2022arf}
Zhang, K., Kolkin, N., Bi, S., Luan, F., Xu, Z., Shechtman, E., Snavely, N.: {ARF}: Artistic radiance fields. In: ECCV. pp. 717--733. Springer (2022)

\bibitem{zhang2020nerf++}
Zhang, K., Riegler, G., Snavely, N., Koltun, V.: {NeRF++}: Analyzing and improving neural radiance fields. arXiv preprint arXiv:2010.07492  (2020)

\bibitem{zhang2018unreasonable}
Zhang, R., Isola, P., Efros, A.A., Shechtman, E., Wang, O.: The unreasonable effectiveness of deep features as a perceptual metric. In: CVPR. pp. 586--595 (2018)

\bibitem{zhang2023ref}
Zhang, Y., He, Z., Xing, J., Yao, X., Jia, J.: {Ref-NPR}: Reference-based non-photorealistic radiance fields for controllable scene stylization. In: CVPR. pp. 4242--4251 (2023)

\end{thebibliography}
\clearpage

\appendix

\noindent{\LARGE \bfseries Supplementary Material}
\vspace{1em}
In this document, we provide supplementary material that additionally supports the claims of the manuscript. The supplementary material is organized as follows: We present additional ablation study results in Sec. \ref{sec:Add_ablation}. Sec. \ref{sec:Add_rets} provides additional visual results.

\section{Additional Ablation Study Results}
\label{sec:Add_ablation}
\begin{figure}[hb]
    \centering
    \includegraphics[width=1\linewidth]{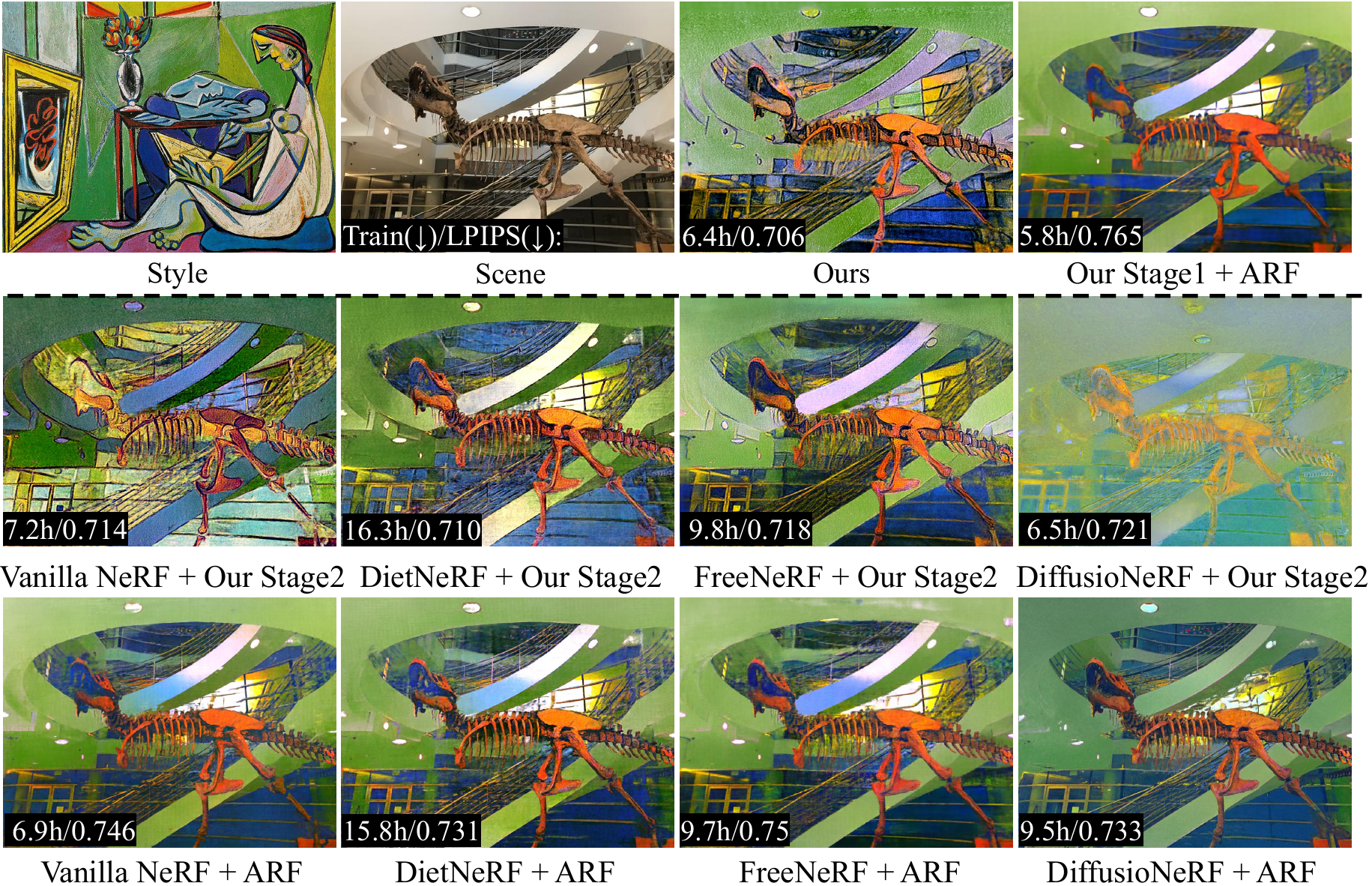}
    \caption{Additional ablation results on coarse-to-fine framework. The top row images are reference style image, the scene and the stylization results of our method and ARF-based fune-tuning of our coarse radiance field. The top images in the bottom group are generated by replacing our coarse radiance field with different fine radiance fields (vanilla NeRF \cite{mildenhall2020nerf}, DietNeRF \cite{jain2021putting}, FreeNeRF \cite{yang2023freenerf}, DiffusioNeRF \cite{wynn2023diffusionerf}). The bottom row images in the bottom group are rendered by directly fine-tuning different pre-trained fine radiance fields with ARF.}
\label{fig:ablation_sup}
\end{figure}

In Fig. \ref{fig:ablation_sup}, we show additional ablation results on coarse-to-fine framework. Four variations replace our first stage with finer scene representations reconstructed from vanilla NeRF \cite{mildenhall2020nerf} and few-shot NeRFs (DietNeRF \cite{jain2021putting}, FreeNeRF \cite{yang2023freenerf}, DiffusioNeRF \cite{wynn2023diffusionerf}). The other five variations use ARF\ \cite{zhang2022arf} to fine-tune different pre-trained radiance fields, including our coarse radiance field, vanilla NeRF, DietNeRF, FreeNeRF and DiffusioNeRF.

\section{Additional Visual Results}
\label{sec:Add_rets}

\begin{figure}[h]
    \centering
    \includegraphics[width=0.98\linewidth]{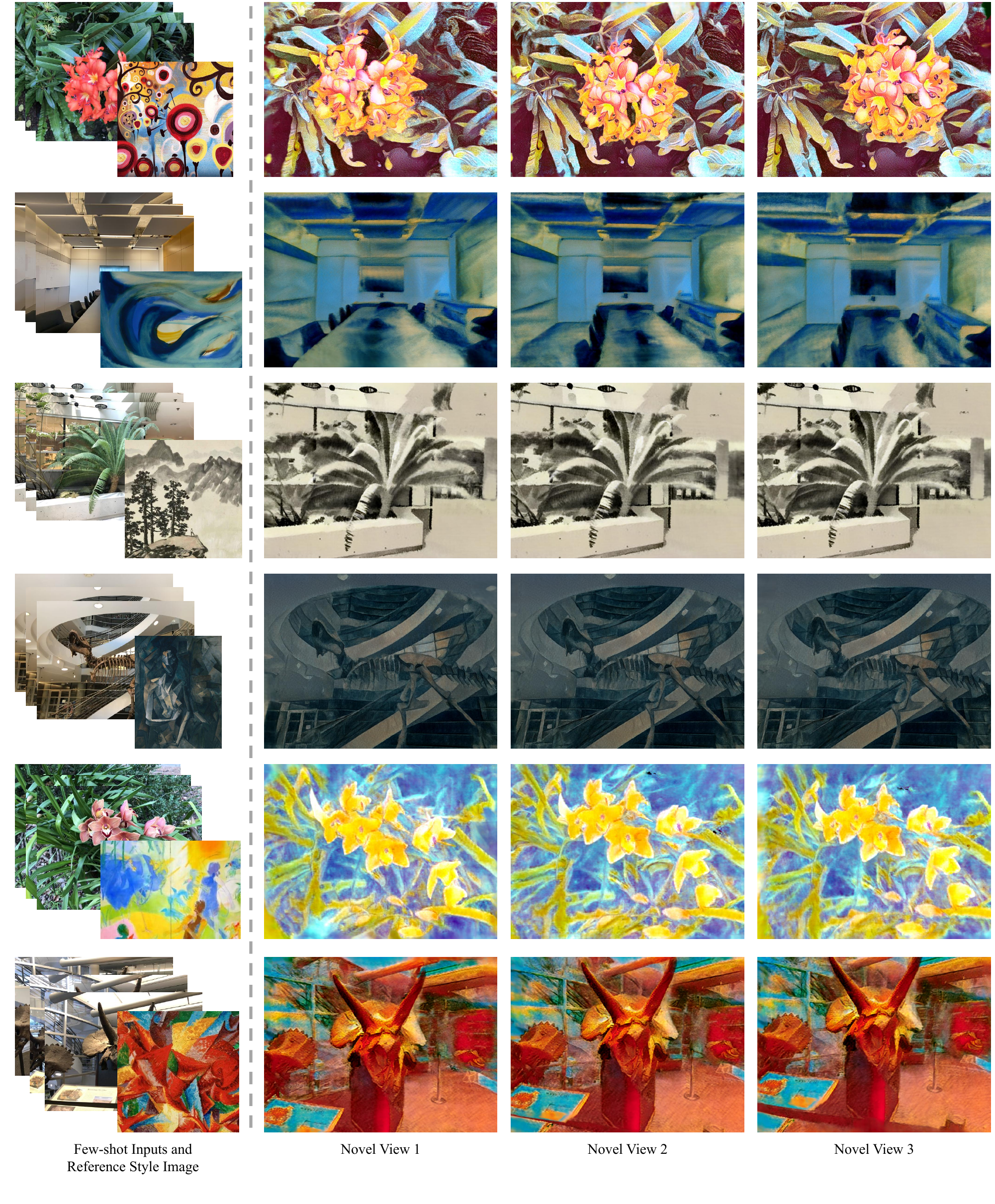}
    \caption{More 3D scene stylization results on LLFF \cite{mildenhall2019local}. The leftmost image in each row contains the few-shot inputs of the scene and its reference style, and the rest are stylized novel views rendered from corresponding scenes. For each scene, we show three novel views rendered from different viewpoints.
    }
\label{fig:figure2_sup}
\end{figure}

\begin{figure}[h]
    \centering
    \includegraphics[width=1\linewidth]{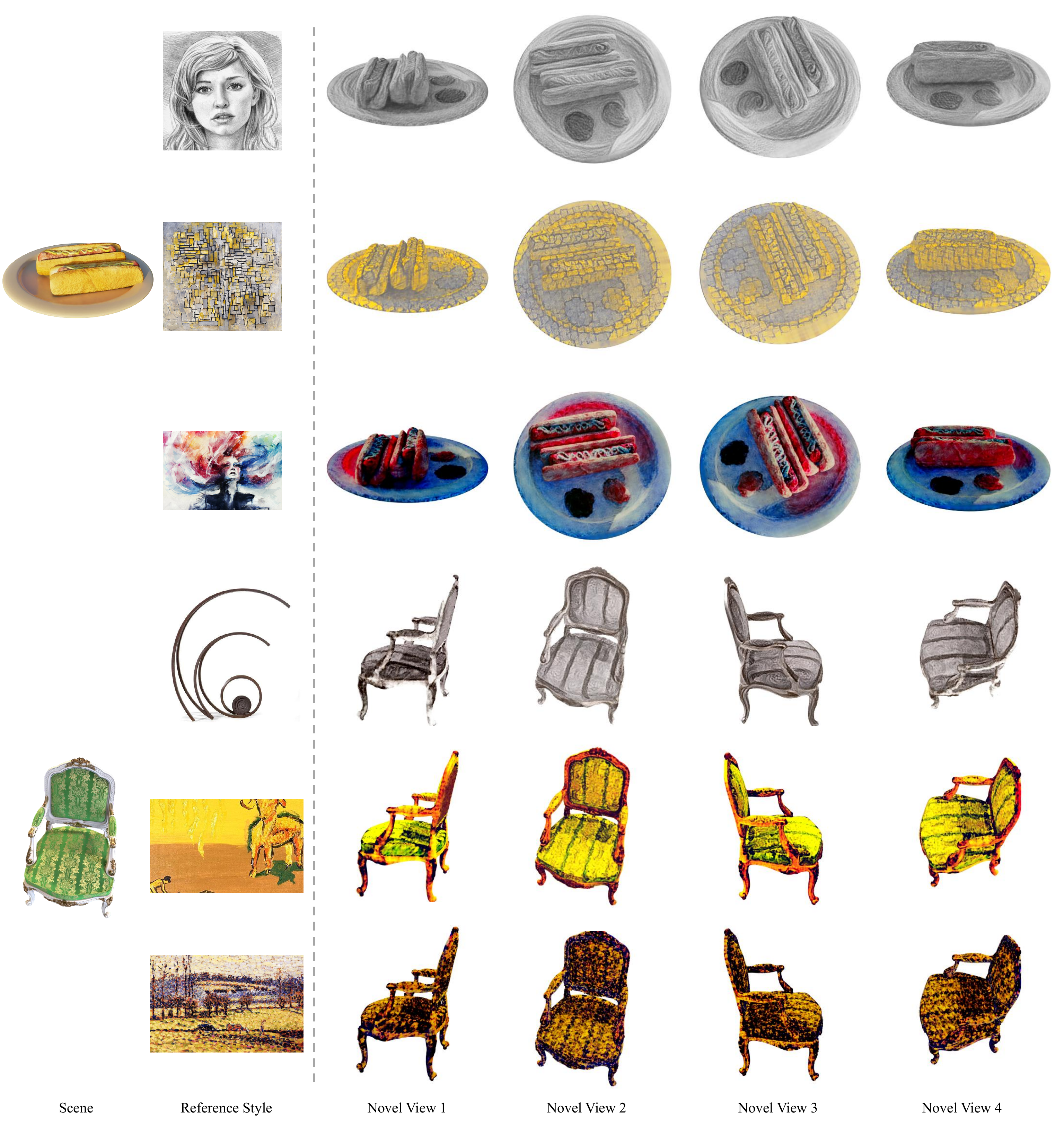}
    \caption{More stylization results of \textit{Hotdog, Chair} with different reference styles on Blender dataset \cite{mildenhall2020nerf}. For each scene, we show four novel views rendered from different viewpoints.
    }
\label{fig:figure3_sup}
\end{figure}

\begin{figure}[h]
    \centering
    \includegraphics[width=1\linewidth]{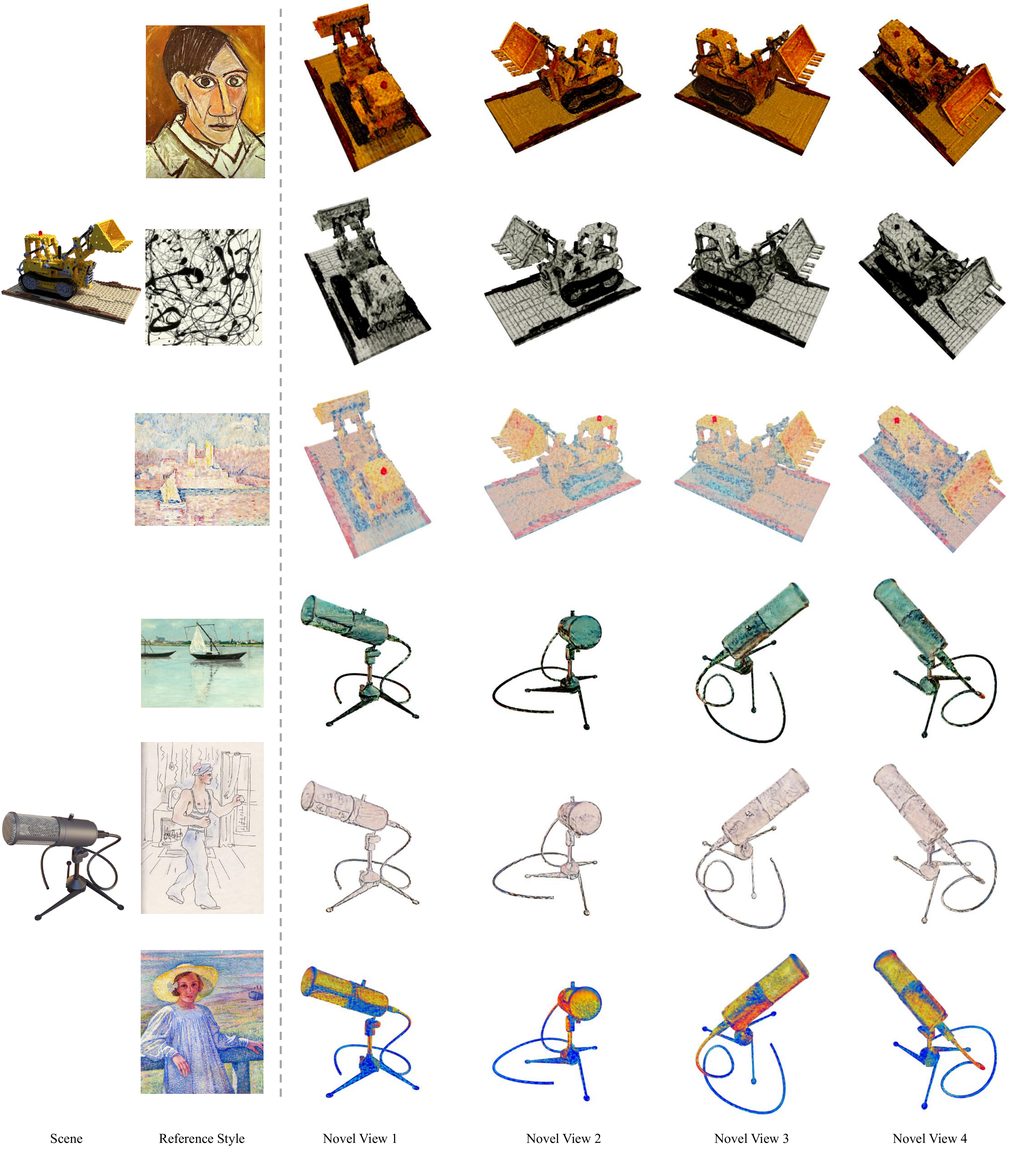}
    \caption{More stylization results of \textit{Lego, Mic} with different reference styles on Blender \cite{mildenhall2020nerf}. For each scene, we show four novel views rendered from different viewpoints.}
\label{fig:figure4_sup}
\end{figure}

To better visualize the multi-view consistency and amazing stylization performance of the proposed method, we present more stylization results of \textit{Flower}, \textit{Room}, \textit{Fern}, \textit{Trex}, \textit{Orchids}, \textit{and Horns} with different style image on LLFF \cite{mildenhall2019local} in Fig. \ref{fig:figure2_sup}. Fig. \ref{fig:figure3_sup} and \ref{fig:figure4_sup} show more 3D scene stylization results on Blender dataset \cite{mildenhall2020nerf}. From these results, we observe that our method achieves the best multi-view consistency and can handle a wide variety of reference styles. We also recommend readers watch the provided videos for a better comparison.



%
%

\end{document}